\documentclass{article}

\usepackage{PRIMEarxiv}

\usepackage{graphicx,amsmath,url,amsfonts,bm,subcaption,booktabs,nccmath}
\usepackage[mathscr]{eucal}
\usepackage{hyperref,cleveref}
\hypersetup{colorlinks=true,linkcolor=[rgb]{0.1,0.3,0.9},urlcolor=[rgb]{0.1,0.3,0.9},citecolor=[rgb]{0.1,0.3,0.9}}

\title{Learning spatiotemporal features from incomplete data for traffic flow prediction using hybrid deep neural networks}

\author{
Mehdi Mehdipour Ghazi \\
Faculty of Mechanics, Electrical and Computer \\
Science and Research Branch IAU \\
Tehran, Iran \\
\texttt{mehdi.mehdipour@srbiau.ac.ir} \\
\And
Amin Ramezani \\
Faculty of Electrical and Computer Engineering \\
Tarbiat Modares University \\
Tehran, Iran \\
\texttt{ramezani@modares.ac.ir} \\
\And
Mehdi Siahi \\
Faculty of Mechanics, Electrical and Computer \\
Science and Research Branch IAU \\
Tehran, Iran \\
\texttt{mehdi.siahi@srbiau.ac.ir} \\
\And
Mostafa Mehdipour Ghazi \\
Department of Computer Science \\
University of Copenhagen \\
Copenhagen, Denmark \\
\texttt{ghazi@di.ku.dk} \\
}

\begin{document}
\maketitle

\begin{abstract}
Urban traffic flow prediction using data-driven models can play an important role in route planning and preventing congestion on highways. These methods utilize data collected from traffic recording stations at different timestamps to predict the future status of traffic. Hence, data collection, transmission, storage, and extraction techniques can have a significant impact on the performance of the traffic flow model. On the other hand, a comprehensive database can provide the opportunity for using complex, yet reliable predictive models such as deep learning methods. However, most of these methods have difficulties in handling missing values and outliers. This study focuses on hybrid deep neural networks to predict traffic flow in the California Freeway Performance Measurement System (PeMS) with missing values. The proposed networks are based on a combination of recurrent neural networks (RNNs) to consider the temporal dependencies in the data recorded in each station and convolutional neural networks (CNNs) to take the spatial correlations in the adjacent stations into account. Various architecture configurations with series and parallel connections are considered based on RNNs and CNNs, and several prevalent data imputation techniques are used to examine the robustness of the hybrid networks to missing values. A comprehensive analysis performed on two different datasets from PeMS indicates that the proposed series-parallel hybrid network with the mean imputation technique achieves the lowest error in predicting the traffic flow and is robust to missing values up until 21\% missing ratio in both complete and incomplete training data scenarios when applied to an incomplete test data.
\end{abstract}


\section{Introduction}

Accurate forecasting of traffic flow is essential for today's intelligent transportation system (ITS). It is performed based on information obtained from data recorded in different road stations at different timestamps and can help with the reduction of traffic volume and accident rate by detecting the roads with more congestion and at higher risk of accidents. Therefore, techniques for data collection, storage, and transfer, as well as methods for information extraction play important roles in traffic flow prediction and many related applications such as routing that require some reliable information about traffic flow.

Traffic flow prediction models can be divided into two categories of analytical and data-driven. The analytical models are developed based on accurate physical laws and can be classified into microscopic, macroscopic, and mesoscopic. The microscopic approaches \cite{gipps1981behavioural,leclercq2007hybrid,treiber2010three} represent individual characteristics of vehicles such as speed and position; Macroscopic approaches \cite{bagnerini2003multiclass,jin2010kinematic,van2013anisotropy}, on the other hand, describe the traffic flow using group-based characteristics of vehicles such as volume and flow; Finally, the mesoscopic approaches \cite{helbing1997modeling,hoogendoorn2001generic} merge the individual and group-based behaviors of vehicles.

Data-driven models are mathematically developed based on the input-output pairs of real systems and can be classified into parametric, nonparametric, and a combination of both. The model structure in the parametric models is prespecified with a finite number of parameters for fitting the data. For example, the study in \cite{kumar2015short} used autoregressive (AR) models for predicting short-term traffic flow considering the existing seasonality or periodicity among the data with a stationary assumption, i.e., constant mean and variance for time series. Moreover, Kalman filter \cite{guo2014adaptive} and Gaussian mixture model (GMM) \cite{jin2013short}, among other parametric methods, were applied to traffic flow prediction. Despite their simplicity, the parametric methods are not flexible due to their assumptions and cannot effectively describe the behavior of traffic data which is random, nonlinear, and spatially interconnected \cite{vlahogianni2014short}. On the other hand, the stationary assumption cannot hold for long-term predictions as the traffic flow can be affected by weather conditions and events, and be different for weekdays and weekends.

Nonparametric models are flexible and can describe the behavior of data with a large number of parameters. Still, they require more data for learning and often solve a nonconvex problem to find the local optimum. For instance, the research study in \cite{chan2011neural} used artificial neural networks (ANNs) and that in \cite{wu2015short} applied k-nearest neighbor (KNN) regression to predict the traffic flow. Also, support vector regression (SVR) \cite{wang2013short} was used to predict the traffic speed. These models can converge to the optimal solution, but one needs to apply a proper kernel function. To further improve the prediction performance, a combination of several parametric/nonparametric methods can be utilized to obtain more adaptive and flexible models. However, they are more complex and computationally expensive and not able to fully describe (spatially or temporally) big traffic data.

With the rapid development of deep learning networks and their success in different applications \cite{krizhevsky2012imagenet}, they have been attracting interest in transportation and traffic fields \cite{huang2014deep}. Compared to shallow ANNs which use a few layers, deep learning approaches benefit from many cascaded layers that can learn deep representations from large amounts of data and high-dimensional features. For instance, stacked autoencoders (SAEs) \cite{lv2014traffic} were used for the first time to learn deep representation from the utilized data to predict the traffic flow. To improve the prediction accuracy, an SAE based on the Lundberg-Markard model was used \cite{yang2016optimized} as an optimized architecture using unsupervised learning. However, these methods employ fully-connected networks in their architecture that cannot describe the spatiotemporal characteristics of traffic data.

To address temporal dependencies in traffic data, state-of-the-art recurrent neural networks (RNNs) such as long-short term memories (LSTMs) \cite{hochreiter1997long} and gated recurrent units (GRUs) \cite{cho2014learning} were used to predict the volume and speed of traffic. Accordingly, an LSTM network \cite{ma2015long} was applied to effectively predict traffic speed. Likewise, the superiority of LSTM and GRU networks in predicting the traffic flow over AR models was represented in \cite{fu2016using}. However, these models did not take spatial correlations among data into account.

To consider the spatial correlations between different stations, convolutional neural networks (CNNs) have been used either separately or in combination with RNNs. For example, CNNs were applied to time series of urban population flow \cite{zhang2016dnn} at different places assuming them as video channels. More effectively, hybrid networks were proposed based on RNNs and CNNs \cite{yu2017spatiotemporal,wu2018hybrid} to take full advantage of available spatial and temporal features and improve the prediction performance. Still, the traffic data used in these studies are assumed ideal which has no missing or erroneous values.

Real-world traffic data can include missing information or outliers for various reasons, which can lead to unpredictability or a decline in prediction accuracy. Several data imputation methods have been proposed for traffic data in the literature to retrieve and fill missing values \cite{tak2016data,bae2018missing}. However, their performance in predicting traffic flow using state-of-the-art deep learning methods has not thoroughly been studied. As a result, in this work, historical data acquired from the California Freeway Performance Measurement System (PeMS) is studied for traffic flow prediction in real-world scenarios with missing values using hybrid deep learning methods. The proposed hybrid networks are based on LSTM and CNN architectures in parallel-series combinations to efficiently capture spatiotemporal dependencies in the recorded data. Different architecture combinations are examined together with several prevalent missing data imputation techniques. The results show that the proposed series-parallel hybrid network with the mean imputation technique obtains the lowest error in predicting the traffic flow within two different subsets of the PeMS.

The main contribution of this study is three-fold. First, we assess different configurations of hybrid deep neural networks to capture effective spatiotemporal representations for better describing the traffic data compared to the state-of-the-art methods. Second, the learning and predictive power of different deep learning models are studied against various amounts of missing data using different data imputation techniques. Third, the performance of the different deep networks and imputation techniques is validated using different datasets (lanes) of the PeMS.

\section{Methods}

\subsection{LSTM model}

LSTM is one of the most popular types of RNNs which was introduced \cite{hochreiter1997long} to tackle the vanishing gradients problem in long sequences to effectively capture the temporal dependencies. In general, the network works in the same way as the RNN, except that the hidden state is calculated based on a cell state or candidate. The cell consists of four gates called input gate, modulation gate, forget gate, and output gate, and decides to whether update its status or forget changing it. The feedforward calculations of a vanilla LSTM layer can be summarized as

\vspace{-0.1cm}
\begin{flalign*}
& \bm{f}_t = \sigma_g(\bm{W}_f\bm{x}_t + \bm{U}_f\bm{h}_{t-1} + \bm{b}_f) \,, & \\
& \bm{i}_t = \sigma_g(\bm{W}_i\bm{x}_t + \bm{U}_i\bm{h}_{t-1} + \bm{b}_i) \,, & \\
& \bm{z}_t = \sigma_c(\bm{W}_c\bm{x}_t + \bm{U}_c\bm{h}_{t-1} + \bm{b}_c) \,, & \\
& \bm{c}_t = \bm{f}_t\odot \bm{c}_{t-1} + \bm{i}_t\odot\bm{z}_t \,, \\
& \bm{o}_t = \sigma_g(\bm{W}_o\bm{x}_t + \bm{U}_o\bm{h}_{t-1} + \bm{b}_o) \,, & \\
& \bm{h}_t = \bm{o}_t\odot \sigma_c(\bm{c}_t) \,, &
\end{flalign*}

\noindent where $\bm{x}_t$, $\bm{f}_t$, $\bm{i}_t$, $\bm{z}_t$, $\bm{c}_t$, $\bm{o}_t$, and $\bm{h}_t$ are $M$-dimensional vectors of the input signal, forget gate, input gate, modulation gate, cell state, output gate, and hidden output at time $t$, respectively. Moreover, $\bm{W}_f$, $\bm{W}_i$, $\bm{W}_c$, and $\bm{W}_o$ are the input weight matrices of size $p \times q$, where $p$ and $q$ are the number of input and hidden nodes, respectively. In addition, $\bm{U}_f$, $\bm{U}_i$, $\bm{U}_c$, and $\bm{U}_o$ are the hidden weight matrices of size $p \times p$, and $\bm{b}_f$, $\bm{b}_i$, $\bm{b}_c$, and $\bm{b}_o$ are the corresponding $p$-dimensional bias vectors. The nonlinear activation functions $\sigma_g$ and $\sigma_c$ are typically set to the logistic sigmoid and hyperbolic tangent, respectively, and $\odot$ denotes the element-wise product.

\subsection{CNN model}

CNN is a deep learning model that applies convolution operations to process tensor data like images to capture spatial patterns using some kernels \cite{taylor2010convolutional}. A convolution layer is inspired by the visual cortex organization with a restricted visual field and designed to extract local features from overlapping areas based on a receptive field. A feature map is obtained by the element-wise multiplication of the (optimizable) kernel with every overlapping window of the input tensor and summation over the resulting values of each product. Multiple kernels can be learned to filter data and form several feature maps that represent different characteristics of the input maps. In contrast to fully-connected networks (FCNs), where every node in one layer is connected to all nodes in the next layer, CNNs use fewer parameters to take the hierarchical patterns in data into account. Hence, they can be seen as regularized networks that can by design avoid overfitting using simpler patterns described by their filters.

\subsection{Hybrid model}

The utilized hybrid network is based on a combination of LSTMs and CNNs and is inspired by the model proposed in \cite{wu2018hybrid}. Assume that $\bm{S}$ is a near-term traffic flow array of size $p \times n$ obtained from $p$ stations/detectors at $n$ time points until the prediction time $t$. The aim is to train a hybrid network that can predict the future flows in each station with a prediction horizon of $h$ using data structured as

\vspace{-0.1cm}
\begin{flalign*}
& \bm{S} = 
\begin{bmatrix}
s_1(t-n) & s_1(t-n+1) & \dots & s_1(t-1) \\
s_2(t-n) & s_2(t-n+1) & \dots & s_2(t-1) \\
\vdots & \vdots & \ddots & \vdots \\
s_p(t-n) & s_p(t-n+1) & \dots & s_p(t-1)
\end{bmatrix}
\,, &
\end{flalign*}

\noindent where $s_1$ and $s_p$ indicate the traffic flows at the upstream and downstream stations, respectively.

Since there is a periodicity in the traffic data induced by the flow pattern difference between weekdays and weekends or holidays, the network needs to be trained using data from previous days and weeks to be able to optimally model such behaviors and accurately predict the future. Therefore, beside the abovementioned near-term traffic flows, data from similar time intervals of the previous days and weeks can be utilized as

\vspace{-0.1cm}
\begin{flalign*}
& \bm{S}_d = 
\begin{bmatrix}
s_1(t_d-n_d) & s_1(t_d-n_d+1) & \dots & s_1(t_d+n_d+h-1) \\
s_2(t_d-n_d) & s_2(t_d-n_d+1) & \dots & s_2(t_d+n_d+h-1) \\
\vdots & \vdots & \ddots & \vdots \\
s_p(t_d-n_d) & s_p(t_d-n_d+1) & \dots & s_p(t_d+n_d+h-1)
\end{bmatrix}
\,, & \\
& \bm{S}_w = 
\begin{bmatrix}
s_1(t_w-n_w) & s_1(t_w-n_w+1) & \dots & s_1(t_w+n_w+h-1) \\
s_2(t_w-n_w) & s_2(t_w-n_w+1) & \dots & s_2(t_w+n_w+h-1) \\
\vdots & \vdots & \ddots & \vdots \\
s_p(t_w-n_w) & s_p(t_w-n_w+1) & \dots & s_p(t_w+n_w+h-1)
\end{bmatrix}
\,, &
\end{flalign*}

\noindent where $\bm{S}_d$ and $\bm{S}_w$ indicate the traffic flow arrays of size $p \times (2n_d+h)$ and $p \times (2n_w+h)$ obtained in the previous day and week, respectively, and $t_d$ and $t_w$ are the corresponding times of the previous day and week to the current prediction time $t$.

\begin{figure*}[!t]
\centering
\includegraphics[scale=0.5]{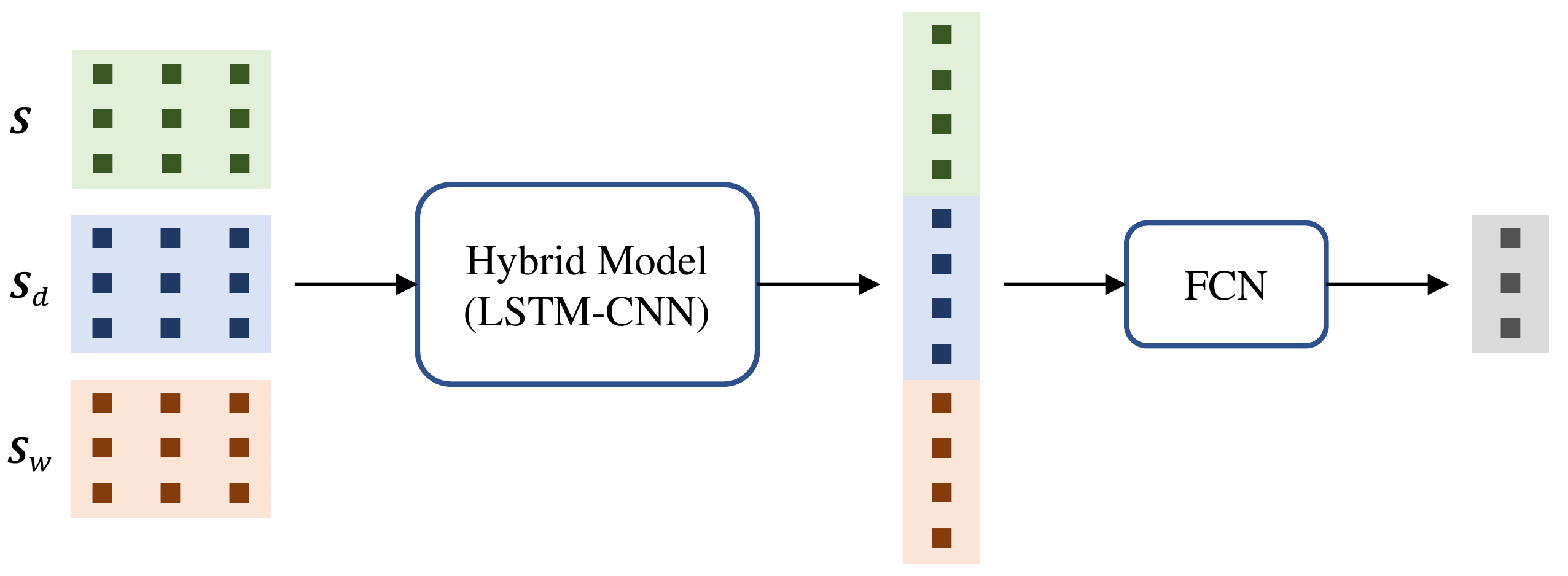}
\caption{A simplified illustration of the utilized prediction model. The CNN and LSTM layers are applied column-wise and row-wise to the traffic flow tensors obtained from the near-term and previous days and weeks data. The extracted representations are then concatenated and summarized using an FCN layer for regression.}
\label{hybrid_fig}
\end{figure*}

In order to extract the spatial features from the traffic data, 1D CNNs can be applied to each column (per time step) of the defined traffic flow arrays with rows (stations) sorted from upstream to downstream. Besides, LSTMs can be applied to each row (per station) of the traffic flow arrays from past to present to capture both the long-term and short-term dependencies of the traffic data. The obtained spatiotemporal representations can then be concatenated and summarized using an FCN layer and used for the regression task. An illustration of the described prediction model is presented in Figure \ref{hybrid_fig}.

\begin{figure*}[!t]
\centering
\includegraphics[scale=0.5]{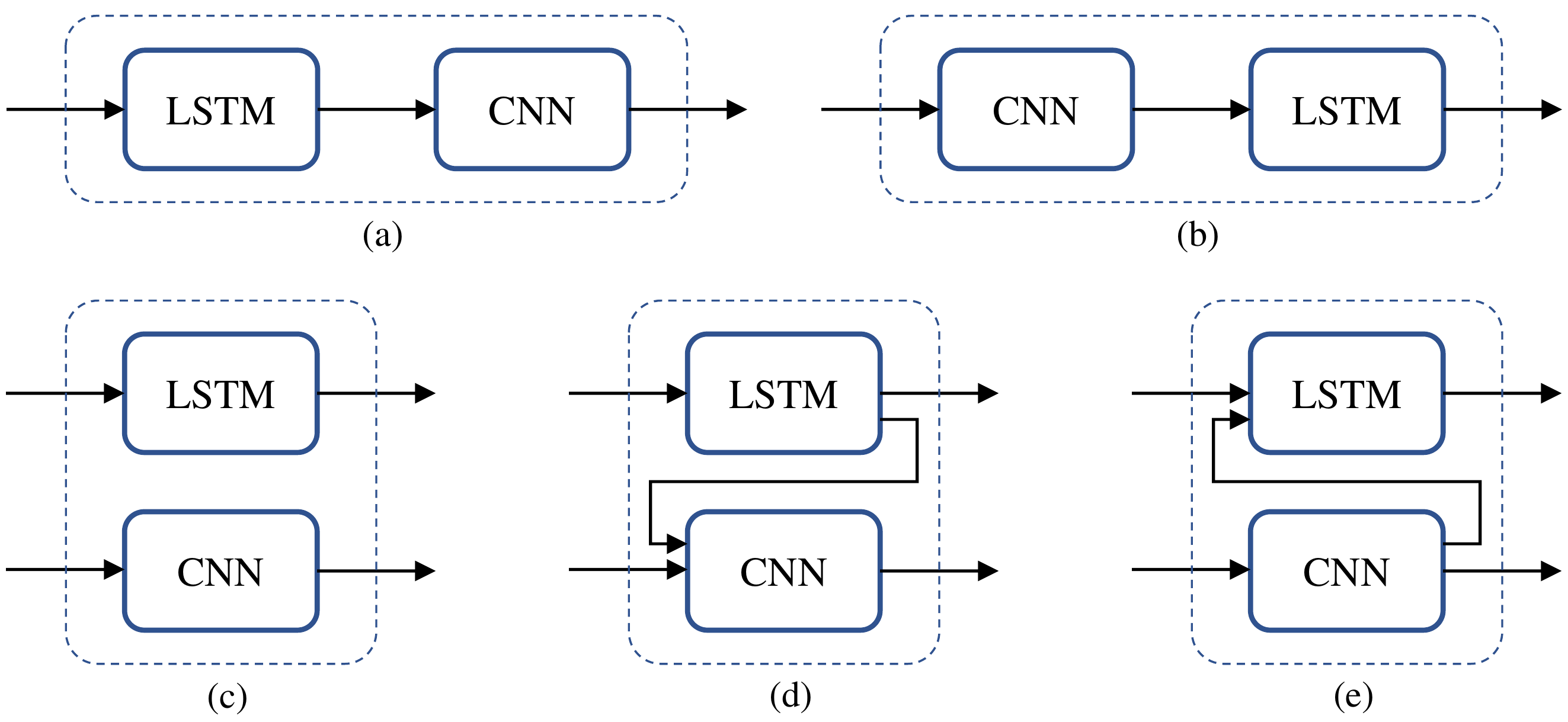}
\caption{Various configurations of the utilized hybrid models for extracting spatiotemporal representations. Subfigures (a) and (b) represent different types of series connections; (c) illustractes a parallel combination; (d) and (e) show different series-parallel connections. Note that multiple LSTM and CNN layers can be used in the combined network.}
\label{combinations_fig}
\end{figure*}

The network proposed in \cite{wu2018hybrid} was based on a parallel combination of the spatial and temporal features extracted individually. However, we investigate different configurations of the hybrid model for spatiotemporal feature extraction. Figure \ref{combinations_fig} shows various forms of the hybrid model based on LSTM-CNN connections. As it can be seen in the figure, there are two types of series-parallel connections. These two forms (feedbacks) cannot be achieved simultaneously because the produced loop cannot be solved by deep learning algorithms. It should also be noted that the number of LSTM/CNN layers as well as the number of inputs for considering the periodicity can vary in the combined network.

\section{Experiments and Results}

\subsection{Data}

\begin{figure*}[!t]
\centering
\includegraphics[scale=0.35]{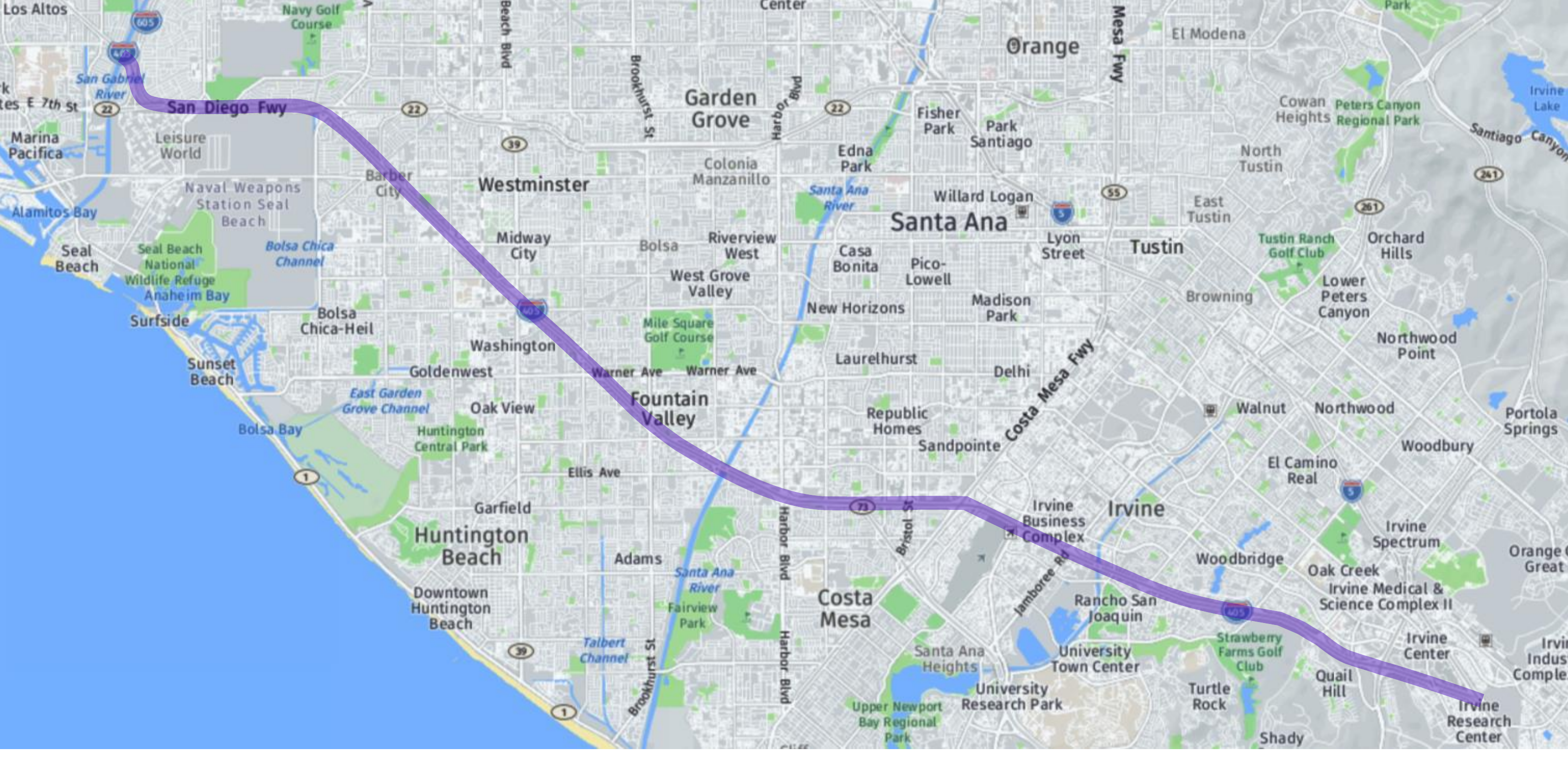}
\caption{Map of the selected zone and road from PeMS. The data obtained from the detectors of zone D12 and road I405-N is used for traffic flow prediction.}
\label{map_fig}
\end{figure*}

Traffic flow prediction data can be acquired from road sensors, entrance-exit detectors, surveillance cameras, radars, or lidars. In this study, the data is obtained from PeMS (\url{http://pems.dot.ca.gov}) that uses inductive loop detectors as road sensors to measure the volume and speed the traffic. The utilized data is collected from a specific area in California freeway with 5-minute intervals, which ensures the efficiency of travel information systems in congested transportation networks \cite{fusco1995use}. The study area is based on the interstate 405 north (I405-N) road in district 12 (D12), as shown in Figure \ref{map_fig}.

\begin{figure*}[!t]
\centering
\begin{subfigure}[t]{0.325\textwidth}
\raisebox{-\height}{\includegraphics[scale=0.38]{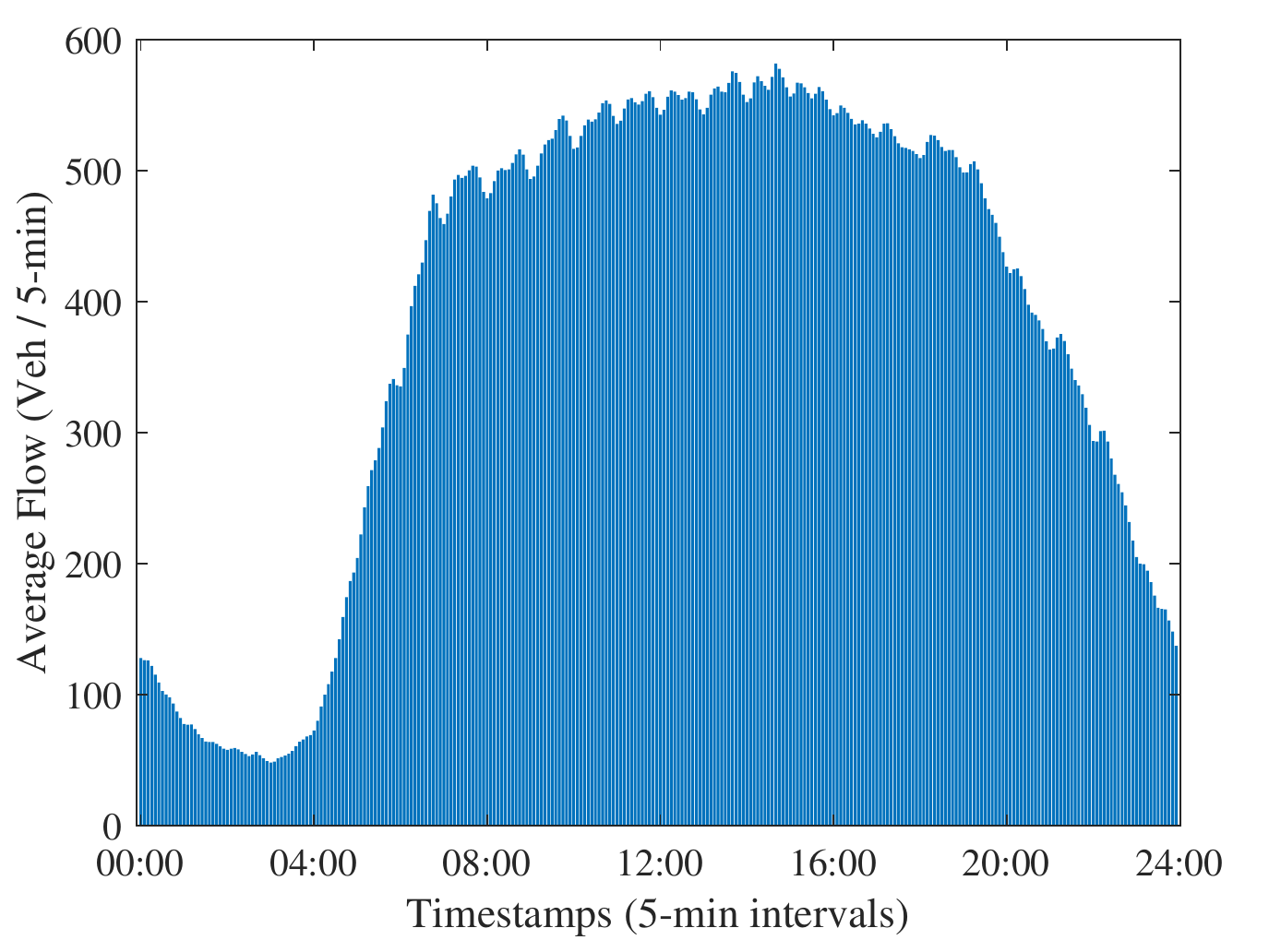}}
\end{subfigure}
\begin{subfigure}[t]{0.325\textwidth}
\raisebox{-\height}{\includegraphics[scale=0.38]{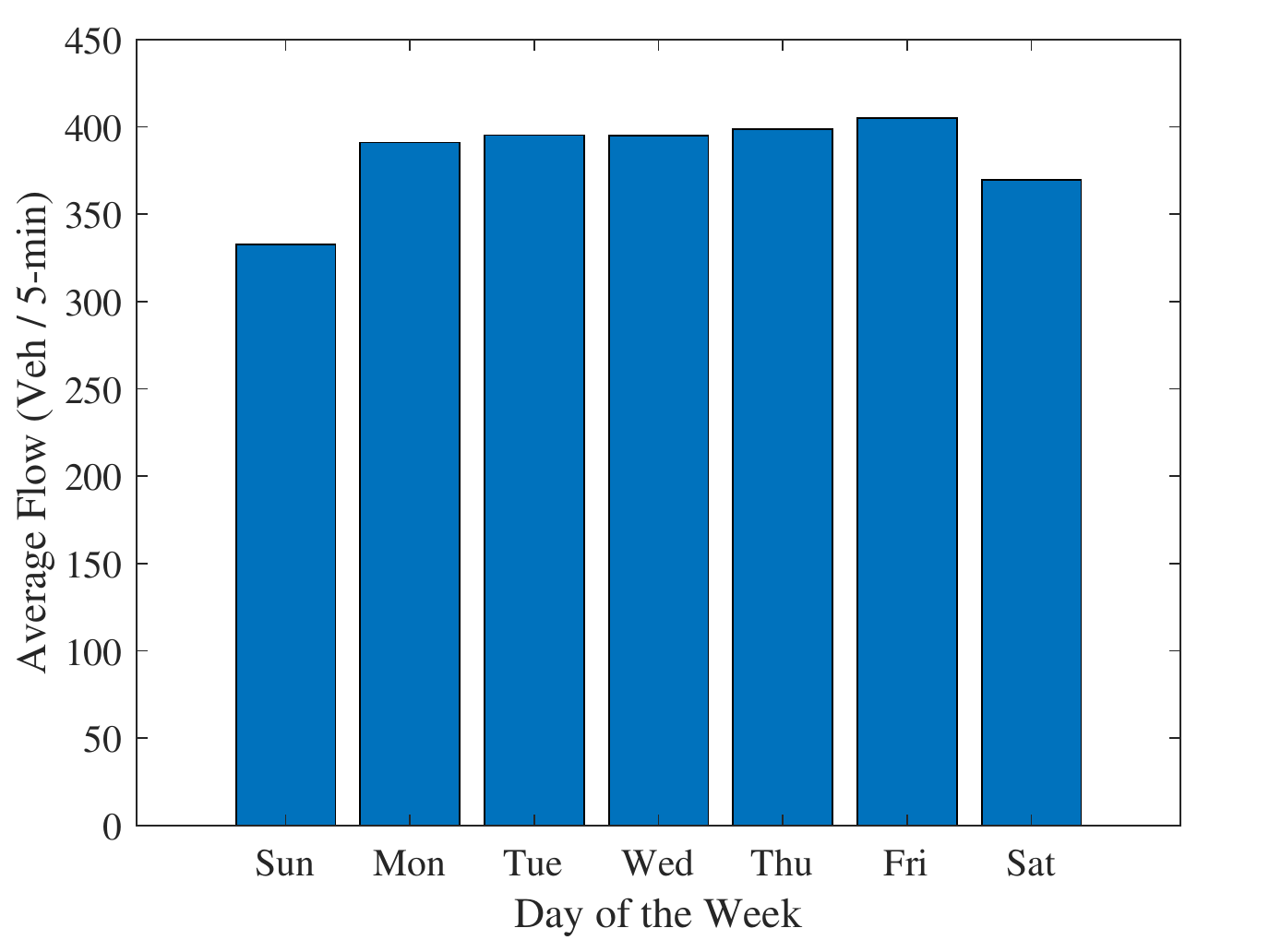}}
\end{subfigure}
\begin{subfigure}[t]{0.325\textwidth}
\raisebox{-\height}{\includegraphics[scale=0.38]{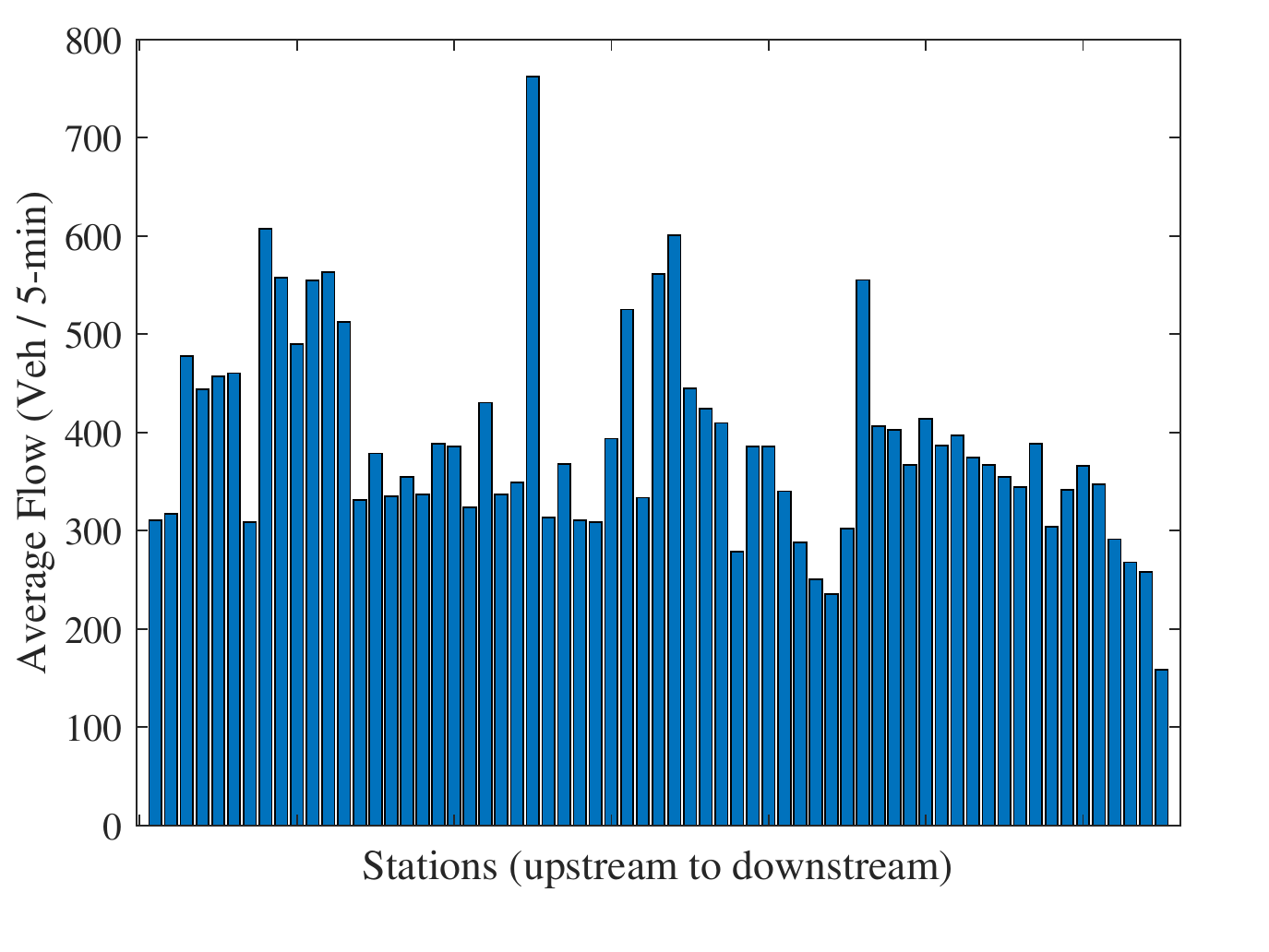}}
\end{subfigure}
\begin{subfigure}[t]{0.325\textwidth}
\raisebox{-\height}{\includegraphics[scale=0.38]{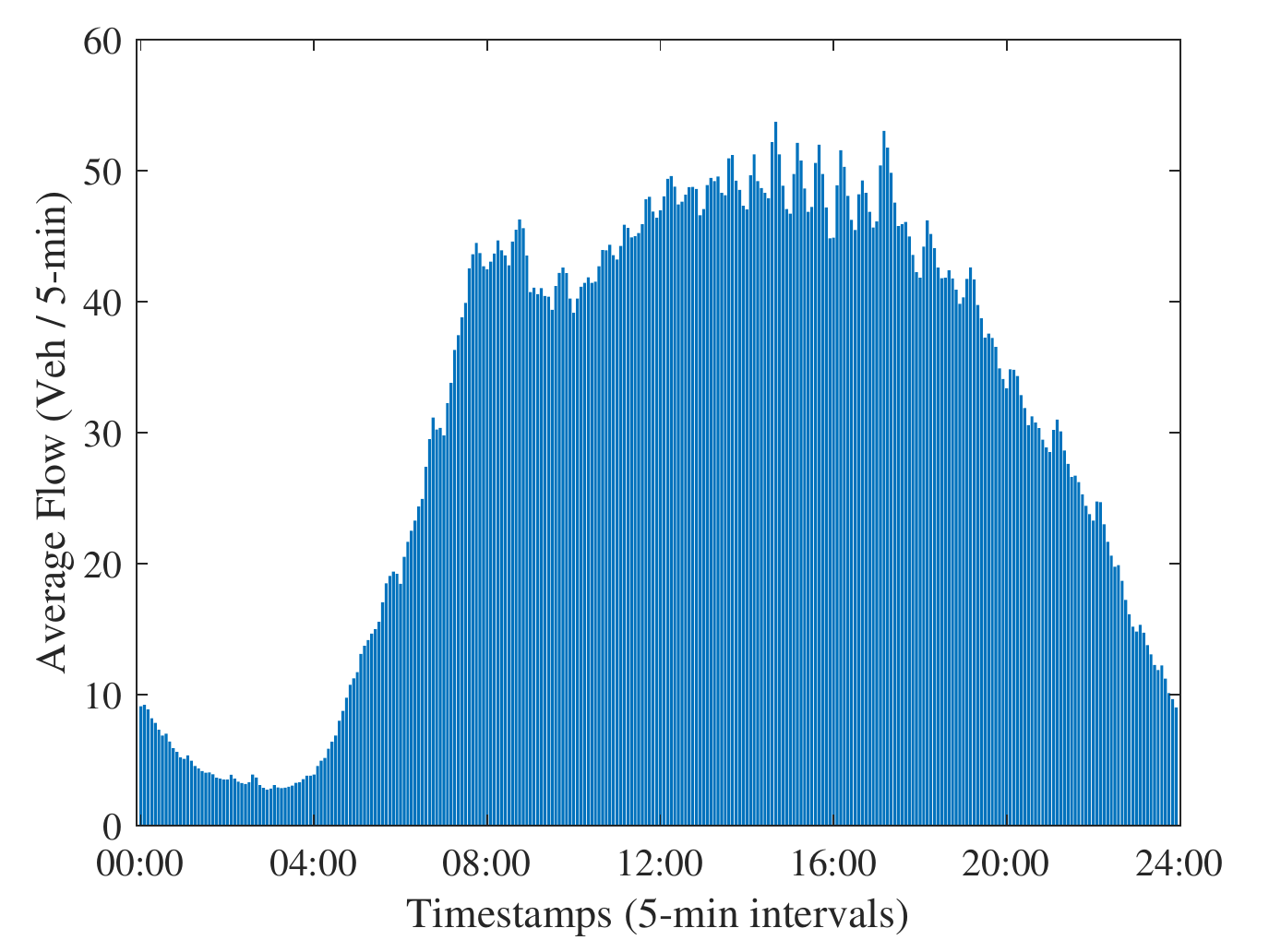}}
\end{subfigure}
\begin{subfigure}[t]{0.325\textwidth}
\raisebox{-\height}{\includegraphics[scale=0.38]{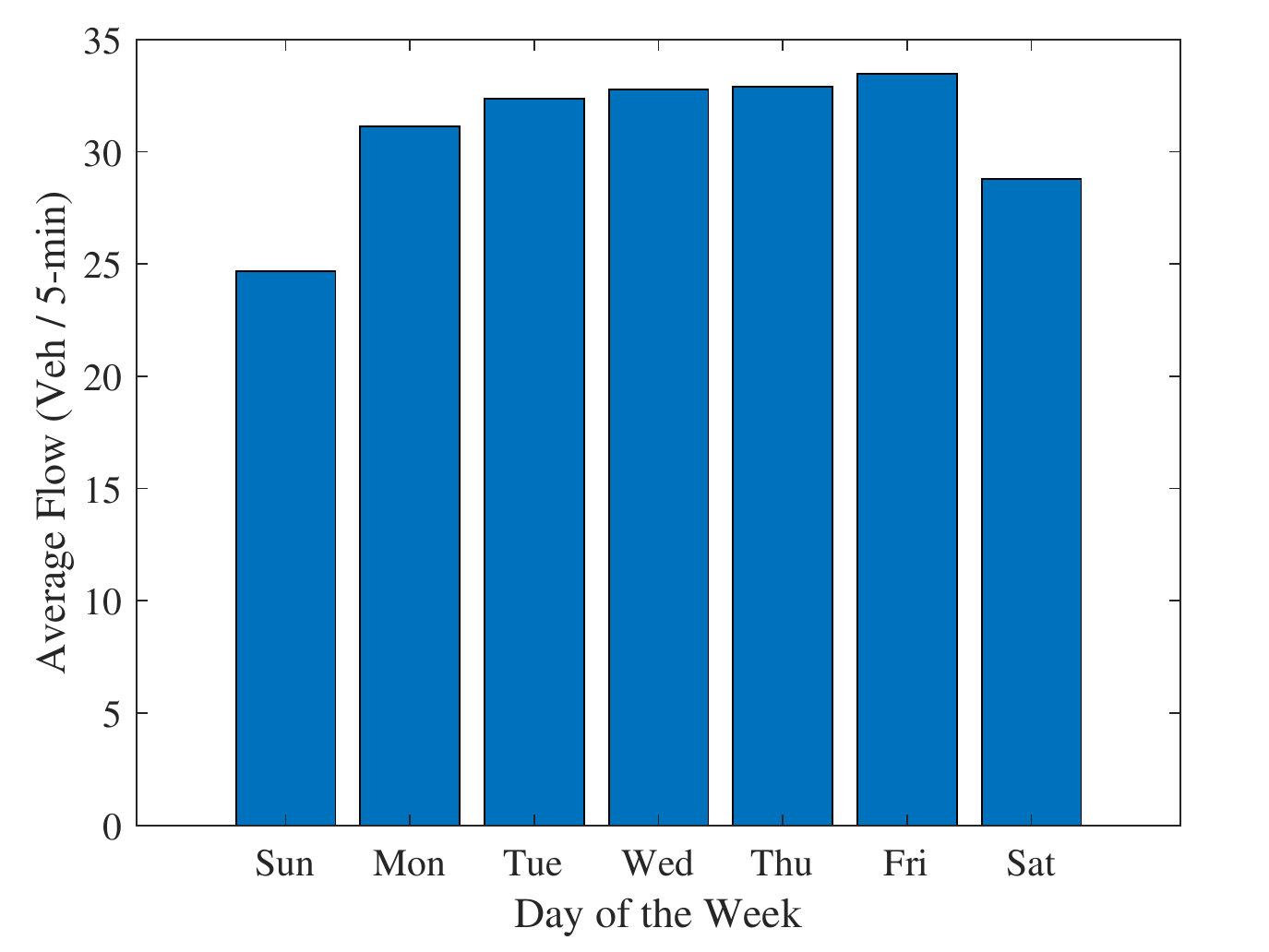}}
\end{subfigure}
\begin{subfigure}[t]{0.325\textwidth}
\raisebox{-\height}{\includegraphics[scale=0.38]{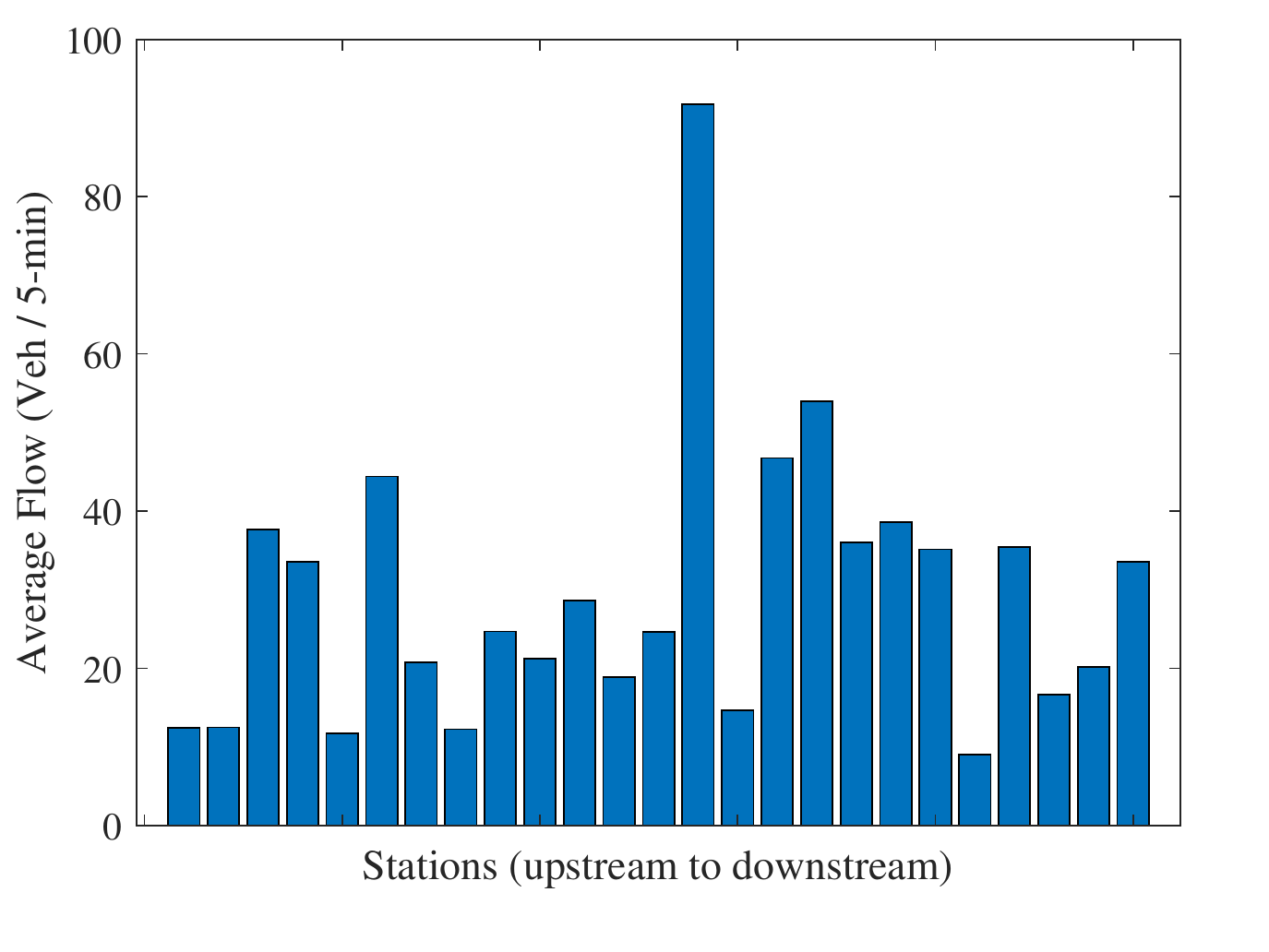}}
\end{subfigure}
\caption{The average traffic flow per timestamp, day, and station. The top and bottom rows are dedicated to the ML and OR lanes of PeMS, respectively.}
\label{stat_fig}
\end{figure*}

We collect daily-basis traffic flow data, each of which contain 288 time points (24 $\times$ 60 min $\div$ 5 min), obtained from the mainline (ML) and on-ramp (OR) lanes of the selected road between 01/01/2019 and 31/01/2020. The OR data includes 25 active detectors and nearly 25\% missing values, while around 7\% of the ML data with 65 active detectors is missing. For our analyses, we sort the detectors based on the given ID of each vehicle detector station (VDS) from upstream to downstream and partition the data into three sets per lane; 80\% for training, 10\% for validation, and 10\% for testing. Figure \ref{stat_fig} shows statistics of the utilized data from the two lanes in terms of the average traffic flow per timestamp, day, and station.

\subsection{Network architectures}

Different deep network architectures based on LSTMs and CNNs are investigated in this work for traffic flow prediction as follows
\begin{itemize}
\item LSTM1: one LSTM unit with $p$ input nodes and $p$ hidden nodes.
\item LSTM2: similar to LSTM1, but with two cascaded LSTM units.
\item LSTM1-S-CNN1: similar to LSTM1, but followed by one CNN layer, as shown in Figure \ref{combinations_fig} (a), with $p$ feature maps.
\item LSTM2-S-CNN3: similar to LSTM1-S-CNN1, but with two cascaded LSTM units and three cascaded CNN layers.
\item CNN1-S-LSTM1: similar to LSTM1-S-CNN1, but in a reversed series order, as shown in Figure \ref{combinations_fig} (b).
\item CNN3-S-LSTM2: similar to LSTM2-S-CNN3, but in a reversed series order.
\item LSTM1-P-CNN1: similar to LSTM1-S-CNN1, but in a parallel connection, as shown in Figure \ref{combinations_fig} (c).
\item LSTM2-P-CNN3: similar to LSTM2-S-CNN3, but in a parallel connection.
\item LSTM1-SP-CNN1: a series-parallel combination of LSTM1-S-CNN1 and LSTM1-P-CNN1, as shown in Figure \ref{combinations_fig} (d).
\item LSTM2-SP-CNN3: similar to LSTM1-SP-CNN1, but with two cascaded LSTM units and three cascaded CNN layers.
\item CNN1-SP-LSTM1: a series-parallel combination of CNN1-S-LSTM1 and LSTM1-P-CNN1, as shown in Figure \ref{combinations_fig} (e).
\item CNN3-SP-LSTM2: similar to CNN1-SP-LSTM1, but with two cascaded LSTM units and three cascaded CNN layers.
\end{itemize}

As suggested in \cite{wu2018hybrid}, the three cascaded CNNs are used with 1D kernel sizes of 4, 3, and 2, respectively. The CNN layers are activated by a rectified linear unit (ReLU), and finally, a fully-connected layer is applied to the concatenated outputs of the hybrid model to reduce the number of output nodes to $p$ for regression. Note that $p$ is set to the number of detectors, i.e., 25 and 65 for the OR and ML lanes, respectively.

\subsection{Imputation methods}

Three different techniques are used for filling missing values as follows
\begin{itemize}
\item Mean: data imputation based on the mean value per station per timestamp obtained across available training data points on different dates.
\item Median: data imputation based on the median value per station per timestamp obtained across available training data points on different dates.
\item Interpolation: data imputation based on linear interpolation of the available training data points on different dates per station per timestamp.
\end{itemize}

\subsection{Evaluation metrics}

Two common metrics are used for the prediction error analysis; mean absolute error (MAE) and root mean square error (RMSE). The former measures the absolute difference between the actual observation $y_j$ and predicted value $\hat{y}_j$ as

\vspace{-0.1cm}
\begin{flalign*}
& \text{MAE} = \frac{1}{\mathcal{J}} \sum_{j=1}^{\mathcal{J}} \big|y_j - \hat{y}_j\big| \,, &
\end{flalign*}

\noindent where $\mathcal{J}$ is the number of test samples. The latter calculates the square root of the average of squared differences between the actual and predicted values as

\vspace{-0.1cm}
\begin{flalign*}
& \text{RMSE} = \sqrt{\frac{1}{\mathcal{J}} \sum_{j=1}^{\mathcal{J}} \big(y_j - \hat{y}_j\big)^2} \,. &
\end{flalign*}

Compared to RMSE in which errors are squared, MAE can result in smaller values for big differences, and hence, it can suppress the effect of outliers in the predictions \cite{chai2014root}.

\subsection{Optimization setup}

The proposed method can be applied to traffic flow prediction in short-term (minutes), mid-term (hours), or long-term (days) scenarios. In this study, we focus on forecasting traffic flow with a prediction horizon of $h = 9$ (45 min). The training sequence length is set to $n = 21$ and the periodicity time lags are selected as $n_d = n_w = 6$ to ensure that the three flow matrices have the same size $p \times n$. Therefore, considering the daily and weekly periodicities, one can extract up to $253$ ($= 288 - 21 - 9 - 6 + 1$) overlapping sequences per day for training.

Since traffic data only accepts non-negative values, we apply simple thresholding to remove any erroneous values of the samples. Next, we standardize the data to have a zero mean and unit variance per station. Finally, we train the networks for at most 30 epochs using the Adam optimizer \cite{kingma2014adam} with a base learning rate of $10^{-3}$, an L2-norm regularization factor of $10^{-4}$, and a batch size set to the number of samples per day.

\subsection{Results and discussion}

First, we train the aforementioned deep learning models five times on the ML lane dataset with no missing values. The obtained prediction errors are presented in Table \ref{ml_errors_tbl} for both the validation and test sets. As can be seen, the hybrid model LSTM2-SP-CNN3 with a series-parallel connection outperforms the other models in predicting future traffic flows with smaller errors in all cases.

\begin{table*}[t]
\centering
\normalsize
\caption{Prediction performance (mean$\pm$SD) of the regression models trained on the ML lane data in five runs. The best results are highlighted in boldface.}
\label{ml_errors_tbl}
\renewcommand{\arraystretch}{1.25}
\centering
\begin{tabular}{lcccc}
\toprule
 & \multicolumn{2}{c}{Validation} & \multicolumn{2}{c}{Test} \\
Model & MAE & RMSE & MAE & RMSE \\
\bottomrule
LSTM1 & 0.215$\pm$0.008 & 0.295$\pm$0.009 & 0.221$\pm$0.006 & 0.308$\pm$0.007 \\
LSTM2 & 0.220$\pm$0.007 & 0.307$\pm$0.008 & 0.220$\pm$0.007 & 0.316$\pm$0.007 \\
LSTM1-S-CNN1 & 0.219$\pm$0.008 & 0.295$\pm$0.008 & 0.228$\pm$0.007 & 0.312$\pm$0.006 \\
LSTM2-S-CNN3 & 0.228$\pm$0.009 & 0.306$\pm$0.010 & 0.235$\pm$0.008 & 0.323$\pm$0.009 \\
CNN1-S-LSTM1 & 0.236$\pm$0.008 & 0.310$\pm$0.009 & 0.249$\pm$0.007 & 0.335$\pm$0.008 \\
CNN3-S-LSTM2 & 0.248$\pm$0.009 & 0.331$\pm$0.009 & 0.235$\pm$0.008 & 0.328$\pm$0.010 \\
LSTM1-P-CNN1 & 0.239$\pm$0.007 & 0.317$\pm$0.006 & 0.223$\pm$0.006 & 0.303$\pm$0.007 \\
LSTM2-P-CNN3 & 0.214$\pm$0.008 & 0.290$\pm$0.007 & 0.210$\pm$0.007 & 0.291$\pm$0.006 \\
LSTM1-SP-CNN1 & 0.221$\pm$0.007 & 0.300$\pm$0.008 & 0.225$\pm$0.006 & 0.314$\pm$0.007 \\
LSTM2-SP-CNN3 & \textbf{0.208}$\pm$\textbf{0.006} & \textbf{0.282}$\pm$\textbf{0.007} & \textbf{0.207}$\pm$\textbf{0.006} & \textbf{0.290}$\pm$\textbf{0.006} \\
CNN1-SP-LSTM1 & 0.226$\pm$0.007 & 0.301$\pm$0.007 & 0.244$\pm$0.006 & 0.334$\pm$0.007 \\
CNN3-SP-LSTM2 & 0.210$\pm$0.006 & 0.288$\pm$0.007 & 0.234$\pm$0.007 & 0.335$\pm$0.007 \\
\toprule
\end{tabular}
\end{table*}

\begin{figure*}[!t]
\centering
\includegraphics[scale=0.64]{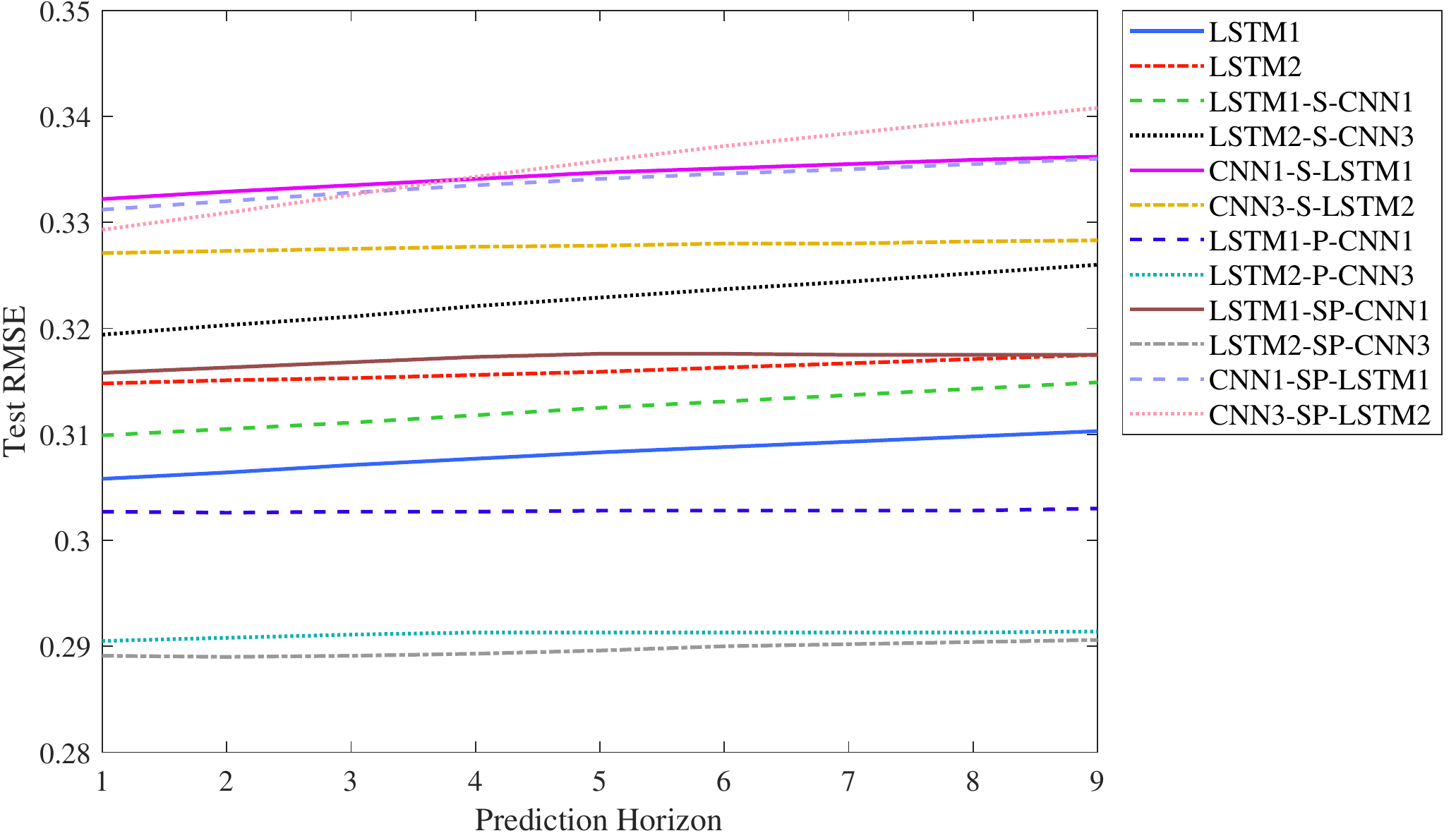}
\caption{The average generalization errors per time step. The errors are obtained using different regression models applied to the ML lane data.}
\label{horizon_fig}
\end{figure*}

Additionally, we inspect the predictive power of the different models within the prediction horizon. Figure \ref{horizon_fig} depicts the average prediction errors of the different models applied to the test set across different runs per time step. As can be seen, the hybrid model LSTM2-SP-CNN3 achieves the best results (lower and stable errors) among the other methods. We select this best model as a baseline for our later analyses.

\begin{figure*}[!t]
\centering
\begin{subfigure}[t]{0.33\textwidth}
\raisebox{-\height}{\includegraphics[scale=0.38]{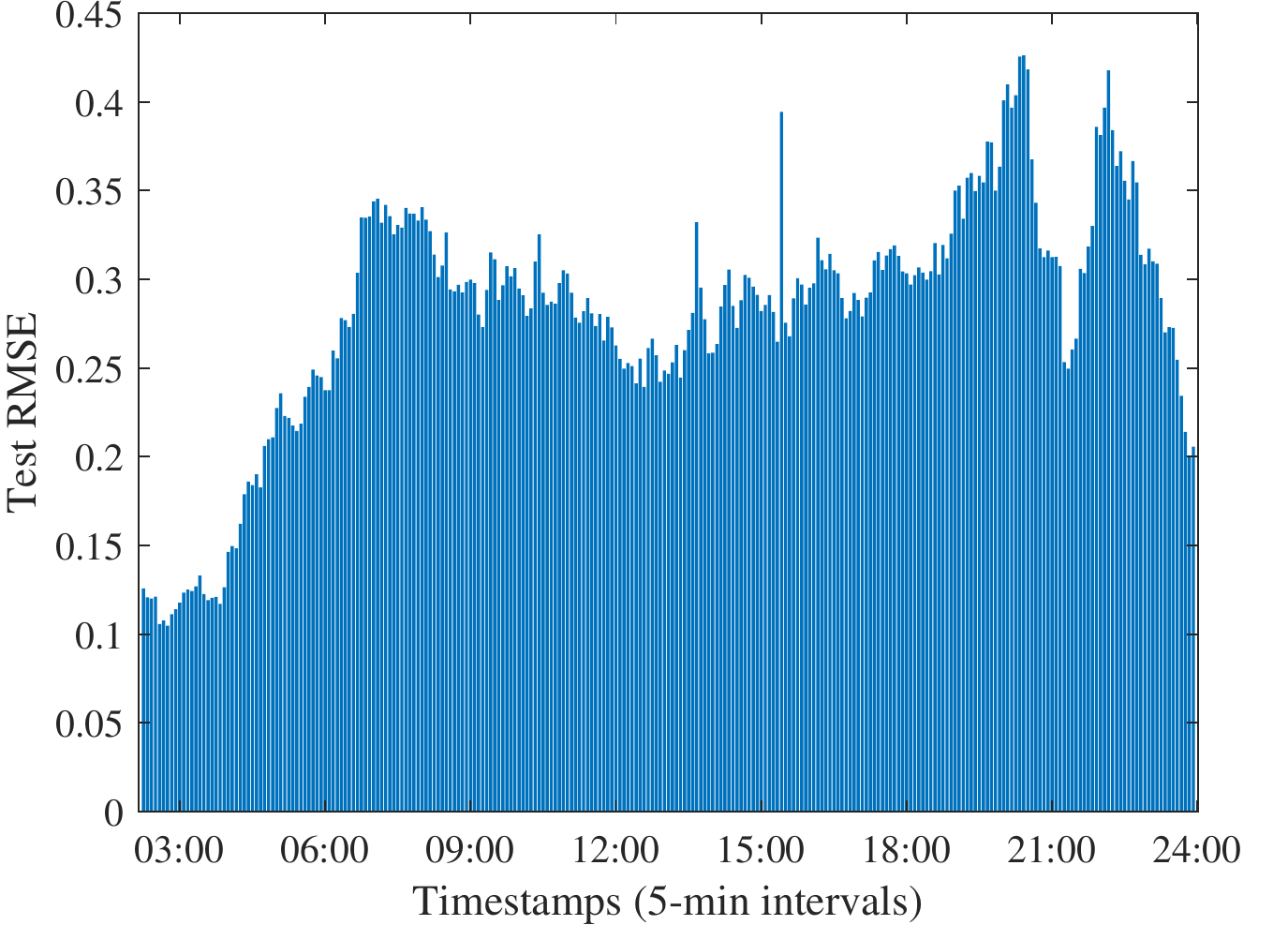}}
\end{subfigure}
\begin{subfigure}[t]{0.325\textwidth}
\raisebox{-\height}{\includegraphics[scale=0.38]{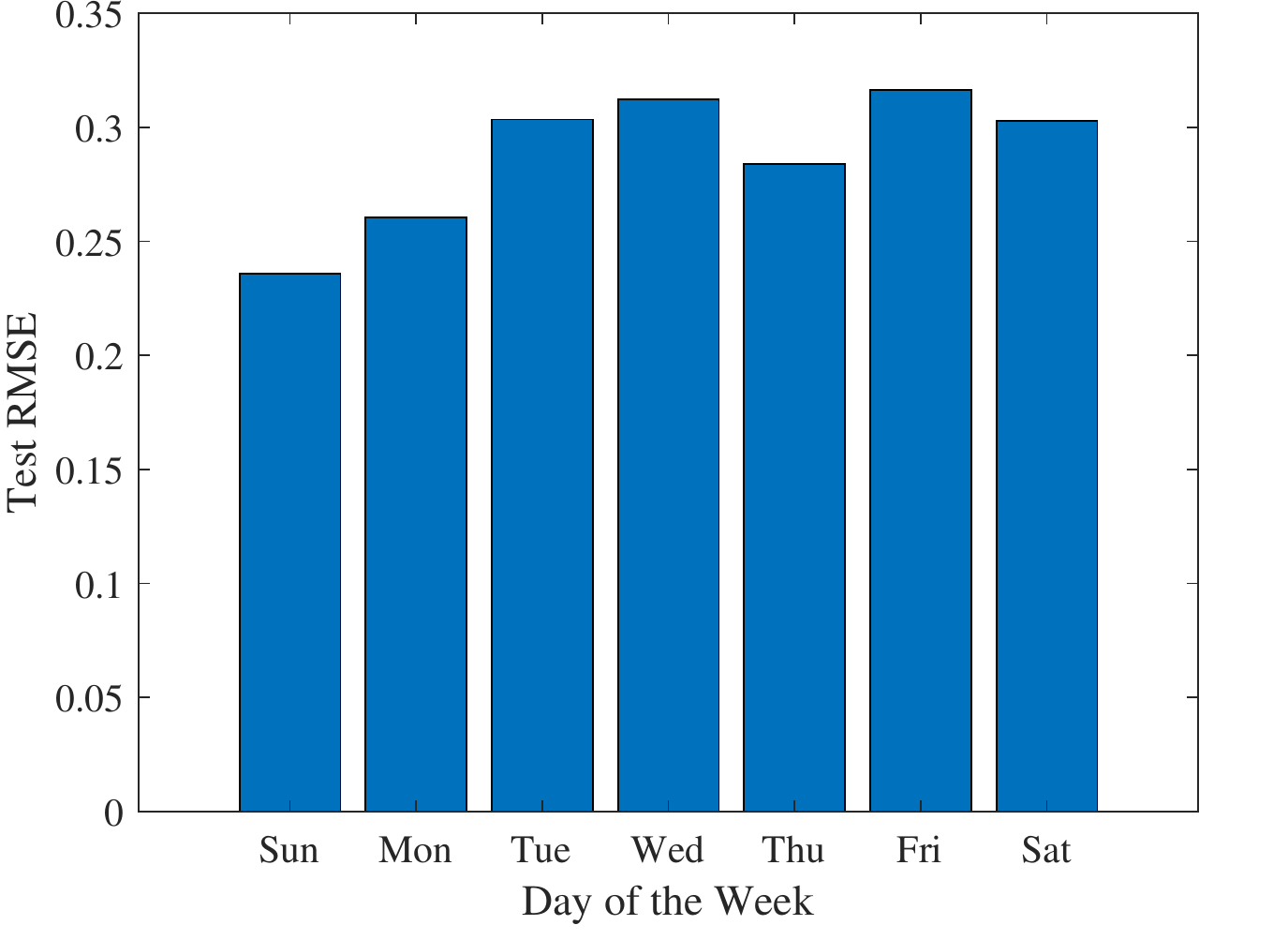}}
\end{subfigure}
\begin{subfigure}[t]{0.325\textwidth}
\raisebox{-\height}{\includegraphics[scale=0.38]{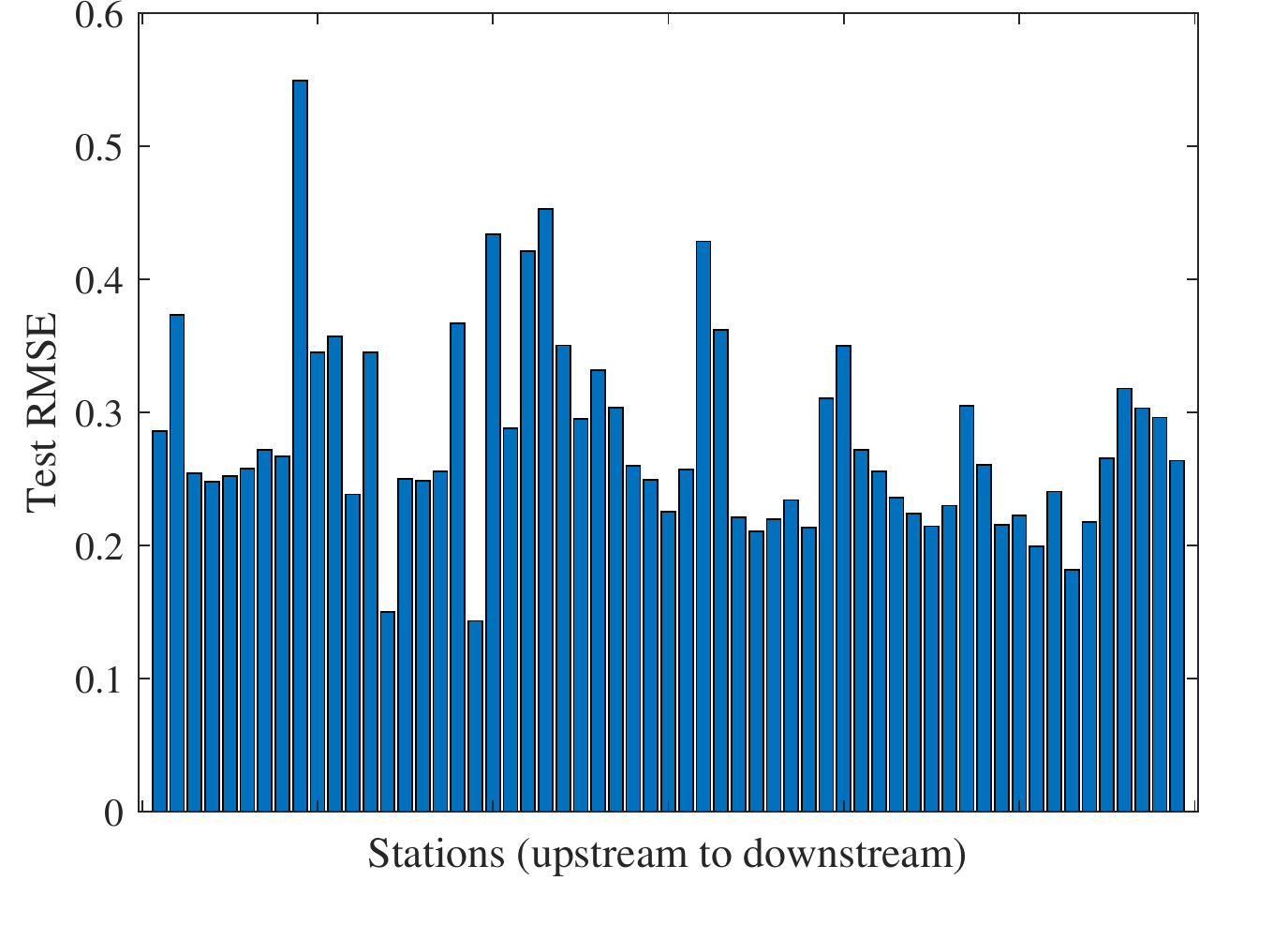}}
\end{subfigure}
\caption{The average generalization errors per timestamp, day, and station. The errors are obtained using the best regression model applied to the ML lane.}
\label{error_fig}
\end{figure*}

Next, we investigate the prediction performance of the best regression model for different stations, timestamps, and days. Figure \ref{error_fig} illustrates the average prediction errors of the test set across different runs per timestamp, day, and station. As it can be deduced, the generalization error difference between different days of the week is not significant. The same behavior is seen between most of the neighboring stations. However, in most of the cases, lower prediction errors are obtained for non-rush hours.

\begin{figure}[!t]
\centering
\includegraphics[scale=0.8]{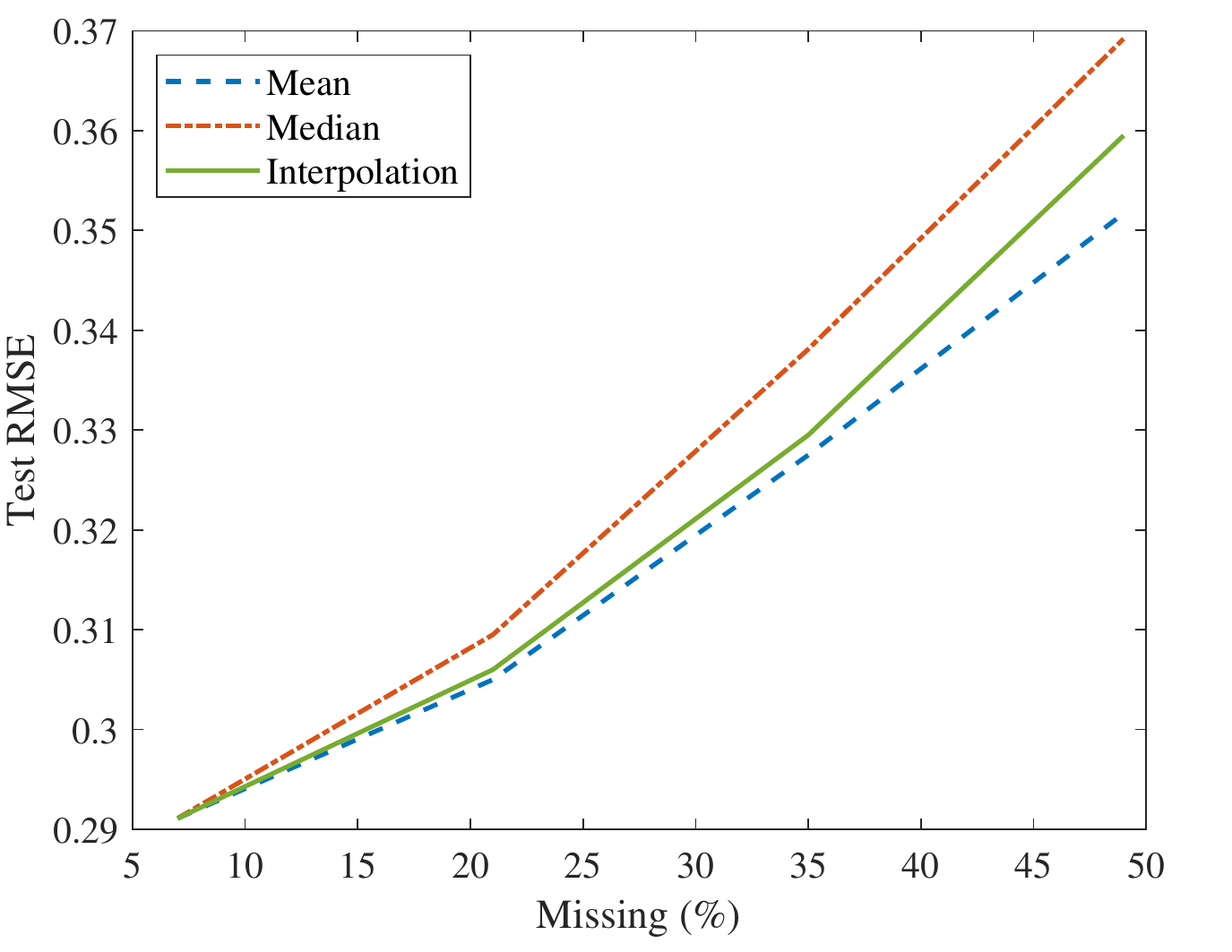}
\caption{The average generalization errors for different test missing ratios and imputation methods using a complete training data. The errors are obtained using the best regression model trained on a ``complete'' data from the ML lane.}
\label{missing_fig2}
\end{figure}

To assess the robustness of the selected model to different amount of missing data, we follow two scenarios. First, we randomly remove some data from the test set and apply the trained model to the incomplete test sets filled by using different imputation methods. Figure \ref{missing_fig2} shows the test prediction errors of the best regression model trained on a complete data for different test missing ratios and imputation methods. It can be seen that missing data imputation using the mean values obtains the lowest generalization errors compared to the alternatives. One possible reason for this result could be that the data does not include many outliers, as the utilization of the median values has not reduced the prediction errors compared to those of the mean values. Moreover, it can be deduced that the model is robust to missing values up until 21\% missing ratio.

\begin{figure}[!t]
\centering
\includegraphics[scale=0.8]{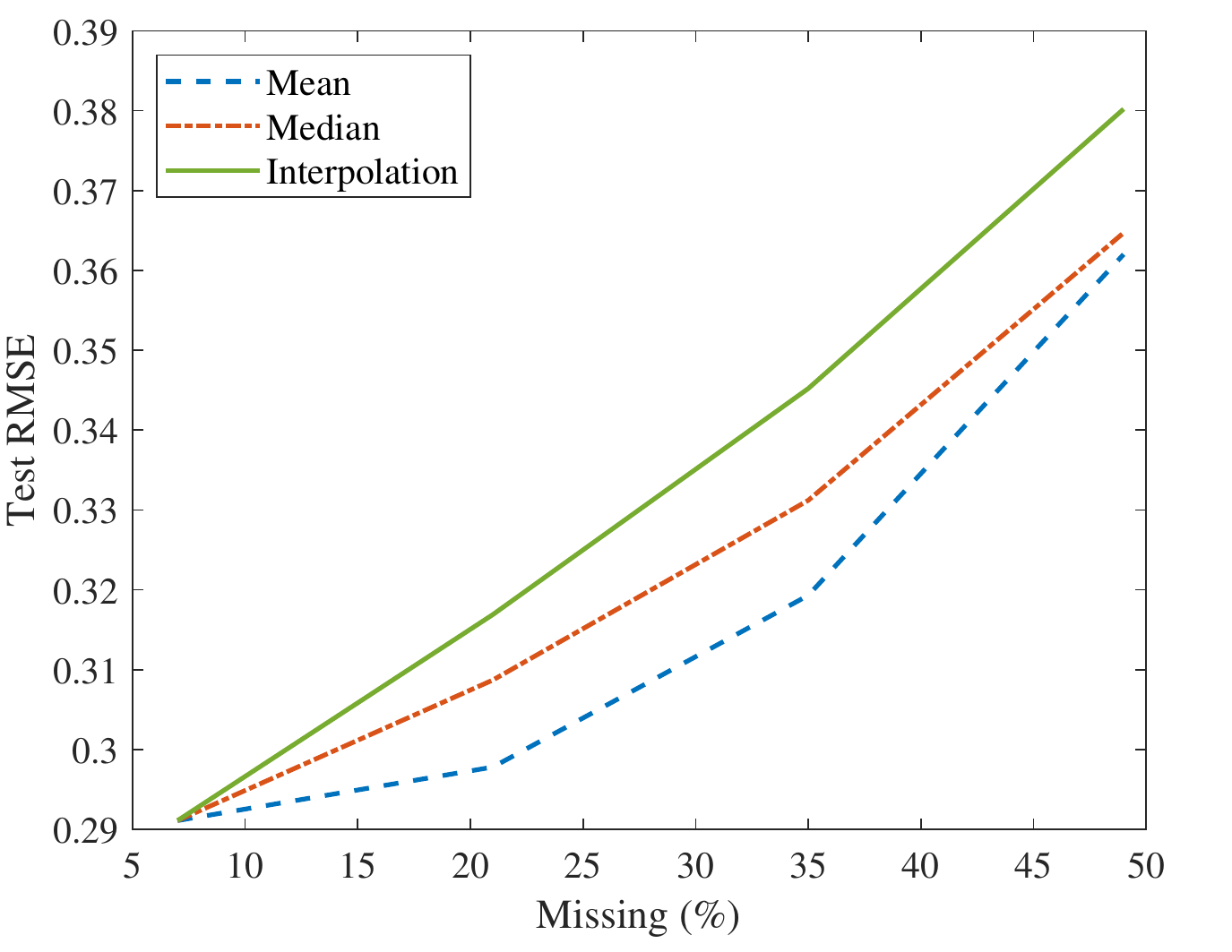}
\caption{The average generalization errors for different test missing ratios and imputation methods using an incomplete training data. The errors are obtained using the best regression model trained on an ``incomplete'' data from the ML lane.}
\label{missing_fig1}
\end{figure}

Next, we randomly remove some data from the entire dataset and train the best hybrid model five times on the incomplete data filled by using different imputation methods. Finally, we apply the trained model to the incomplete test sets filled by using different imputation methods. Figure \ref{missing_fig1} shows the test prediction results of the best regression model trained on an incomplete data for different test missing ratios and imputation methods. As can be seen, the hybrid model trained on data filled by the mean values obtains the lowest generalization errors and its results are robust to missing values up to more than 21\% missing ratio. However, compared to the model trained on a complete data, median imputation achieves better results than those of the linear interpolation.

Finally, to investigate the predictive power of the utilized deep learning models against different datasets, we train the same networks five times on the OR lane dataset with missing values filled by using the mean values and apply them to the validation and test subsets. The obtained prediction errors are presented in Table \ref{or_errors_tbl} for both subsets. As it can be noticed from the table, the same hybrid model LSTM2-SP-CNN3 achieves the best results in most of the cases. However, the error values of the OR lane are higher than those of the ML lane. One reason for the difference could be that there is less amount of data for training, which is due to the fact that the number of OR detectors is smaller than that of ML (25 vs. 65), and a higher amount of missing data is available in the OR data (25\%).

\begin{table*}[t]
\centering
\normalsize
\caption{Prediction performance (mean$\pm$SD) of the regression models trained on the OR lane data in five runs. The best results are highlighted in boldface.}
\label{or_errors_tbl}
\renewcommand{\arraystretch}{1.25}
\centering
\begin{tabular}{lcccc}
\toprule
 & \multicolumn{2}{c}{Validation} & \multicolumn{2}{c}{Test} \\
Model & MAE & RMSE & MAE & RMSE \\
\bottomrule
LSTM1 & 0.339$\pm$0.014 & 0.472$\pm$0.015 & 0.304$\pm$0.013 & 0.420$\pm$0.014 \\
LSTM2 & 0.346$\pm$0.015 & 0.490$\pm$0.016 & 0.303$\pm$0.014 & 0.418$\pm$0.015 \\
LSTM1-S-CNN1 & 0.335$\pm$0.015 & 0.467$\pm$0.015 & 0.302$\pm$0.013 & 0.417$\pm$0.014 \\
LSTM2-S-CNN3 & 0.345$\pm$0.016 & 0.480$\pm$0.016 & 0.312$\pm$0.016 & 0.429$\pm$0.017 \\
CNN1-S-LSTM1 & 0.336$\pm$0.015 & 0.469$\pm$0.015 & 0.300$\pm$0.014 & 0.415$\pm$0.015 \\
CNN3-S-LSTM2 & 0.338$\pm$0.017 & 0.475$\pm$0.016 & 0.302$\pm$0.016 & 0.418$\pm$0.017 \\
LSTM1-P-CNN1 & 0.338$\pm$0.013 & 0.473$\pm$0.013 & 0.303$\pm$0.014 & 0.419$\pm$0.014 \\
LSTM2-P-CNN3 & 0.340$\pm$0.014 & 0.474$\pm$0.014 & 0.303$\pm$0.013 & 0.416$\pm$0.013 \\
LSTM1-SP-CNN1 & 0.335$\pm$0.013 & 0.470$\pm$0.013 & 0.303$\pm$0.013 & 0.416$\pm$0.014 \\
LSTM2-SP-CNN3 & \textbf{0.331}$\pm$0.012 & \textbf{0.465}$\pm$\textbf{0.013} & 0.301$\pm$0.013 & \textbf{0.413}$\pm$\textbf{0.013} \\
CNN1-SP-LSTM1 & 0.338$\pm$\textbf{0.011} & 0.472$\pm$0.013 & \textbf{0.298}$\pm$\textbf{0.012} & 0.415$\pm$0.013 \\
CNN3-SP-LSTM2 & 0.345$\pm$0.014 & 0.483$\pm$0.013 & 0.306$\pm$0.013 & 0.422$\pm$0.014 \\
\toprule
\end{tabular}
\end{table*}

\section{Summary and Conclusion}

In this paper, different series and parallel configurations of hybrid deep neural networks and data imputation techniques were studied for the prediction of traffic flow with different amount of missing data. The proposed network architectures were based on RNNs and CNNs, considering the spatiotemporal dependencies in the data recorded from different stations and timestamps, and the three prevalent imputation techniques were based on the mean, median, and interpolation, assessing the ability of the deep networks in handling missing values.

The results obtained from two different datasets indicated that the series-parallel hybrid network, as shown in Figure \ref{combinations_fig} (d), with the mean imputation technique can achieve the best performance in predicting the traffic flow from an incomplete data. Moreover, the trained models showed good robustness to missing values till more than 21\% missing ratio in both complete/incomplete training data scenarios when applied to an incomplete test data.

We examined many different deep learning architectures to show how different combinations of the networks can capture spatiotemporal representations for modeling traffic data. This study can be generalized to other time-series prediction tasks including spatiotemporal data with missing values or other deep learning architectures such as GRUs and transformers.

\bibliographystyle{unsrt}
\bibliography{references}

\begin{thebibliography}{10}

\bibitem{gipps1981behavioural}
Peter~G Gipps.
\newblock A behavioural car-following model for computer simulation.
\newblock {\em Transportation Research Part B: Methodological}, 15(2):105--111,
  1981.

\bibitem{leclercq2007hybrid}
Ludovic Leclercq.
\newblock Hybrid approaches to the solutions of the
  “lighthill--whitham--richards” model.
\newblock {\em Transportation Research Part B: Methodological}, 41(7):701--709,
  2007.

\bibitem{treiber2010three}
Martin Treiber, Arne Kesting, and Dirk Helbing.
\newblock Three-phase traffic theory and two-phase models with a fundamental
  diagram in the light of empirical stylized facts.
\newblock {\em Transportation Research Part B: Methodological},
  44(8-9):983--1000, 2010.

\bibitem{bagnerini2003multiclass}
Patrizia Bagnerini and Michel Rascle.
\newblock A multiclass homogenized hyperbolic model of traffic flow.
\newblock {\em SIAM Journal on Mathematical Analysis}, 35(4):949--973, 2003.

\bibitem{jin2010kinematic}
Wen-Long Jin.
\newblock A kinematic wave theory of lane-changing traffic flow.
\newblock {\em Transportation Research Part B: Methodological},
  44(8-9):1001--1021, 2010.

\bibitem{van2013anisotropy}
Femke Van Wageningen-Kessels, Bas van't Hof, Serge~P Hoogendoorn, Hans
  Van~Lint, and Kees Vuik.
\newblock Anisotropy in generic multi-class traffic flow models.
\newblock {\em Transportmetrica A: Transport Science}, 9(5):451--472, 2013.

\bibitem{helbing1997modeling}
Dirk Helbing.
\newblock Modeling multi-lane traffic flow with queuing effects.
\newblock {\em Physica A: Statistical Mechanics and its Applications},
  242(1-2):175--194, 1997.

\bibitem{hoogendoorn2001generic}
Serge~P Hoogendoorn and Piet~HL Bovy.
\newblock Generic gas-kinetic traffic systems modeling with applications to
  vehicular traffic flow.
\newblock {\em Transportation Research Part B: Methodological}, 35(4):317--336,
  2001.

\bibitem{kumar2015short}
S~Vasantha Kumar and Lelitha Vanajakshi.
\newblock Short-term traffic flow prediction using seasonal {ARIMA} model with
  limited input data.
\newblock {\em European Transport Research Review}, 7(3):1--9, 2015.

\bibitem{guo2014adaptive}
Jianhua Guo, Wei Huang, and Billy~M Williams.
\newblock Adaptive {Kalman} filter approach for stochastic short-term traffic
  flow rate prediction and uncertainty quantification.
\newblock {\em Transportation Research Part C: Emerging Technologies},
  43:50--64, 2014.

\bibitem{jin2013short}
Sheng Jin, Dian-hai Wang, Cheng Xu, and Dong-fang Ma.
\newblock Short-term traffic safety forecasting using {Gaussian} mixture model
  and {Kalman} filter.
\newblock {\em Journal of Zhejiang University SCIENCE A}, 14(4):231--243, 2013.

\bibitem{vlahogianni2014short}
Eleni~I Vlahogianni, Matthew~G Karlaftis, and John~C Golias.
\newblock Short-term traffic forecasting: {Where} we are and where we’re
  going.
\newblock {\em Transportation Research Part C: Emerging Technologies},
  43:3--19, 2014.

\bibitem{chan2011neural}
Kit~Yan Chan, Tharam~S Dillon, Jaipal Singh, and Elizabeth Chang.
\newblock Neural-network-based models for short-term traffic flow forecasting
  using a hybrid exponential smoothing and {Levenberg}--{Marquardt} algorithm.
\newblock {\em IEEE Transactions on Intelligent Transportation Systems},
  13(2):644--654, 2011.

\bibitem{wu2015short}
Yuankai Wu, Huachun Tan, Jin Peter, Bin Shen, and Bin Ran.
\newblock Short-term traffic flow prediction based on multilinear analysis and
  k-nearest neighbor regression.
\newblock In {\em CICTP 2015}, pages 556--569. 2015.

\bibitem{wang2013short}
Jin Wang and Qixin Shi.
\newblock Short-term traffic speed forecasting hybrid model based on
  chaos--wavelet analysis-support vector machine theory.
\newblock {\em Transportation Research Part C: Emerging Technologies},
  27:219--232, 2013.

\bibitem{krizhevsky2012imagenet}
Alex Krizhevsky, Ilya Sutskever, and Geoffrey~E Hinton.
\newblock {ImageNet} classification with deep convolutional neural networks.
\newblock {\em Advances in Neural Information Processing Systems},
  25:1097--1105, 2012.

\bibitem{huang2014deep}
Wenhao Huang, Guojie Song, Haikun Hong, and Kunqing Xie.
\newblock Deep architecture for traffic flow prediction: deep belief networks
  with multitask learning.
\newblock {\em IEEE Transactions on Intelligent Transportation Systems},
  15(5):2191--2201, 2014.

\bibitem{lv2014traffic}
Yisheng Lv, Yanjie Duan, Wenwen Kang, Zhengxi Li, and Fei-Yue Wang.
\newblock Traffic flow prediction with big data: a deep learning approach.
\newblock {\em IEEE Transactions on Intelligent Transportation Systems},
  16(2):865--873, 2014.

\bibitem{yang2016optimized}
Hao-Fan Yang, Tharam~S Dillon, and Yi-Ping~Phoebe Chen.
\newblock Optimized structure of the traffic flow forecasting model with a deep
  learning approach.
\newblock {\em IEEE Transactions on Neural Networks and Learning Systems},
  28(10):2371--2381, 2016.

\bibitem{hochreiter1997long}
Sepp Hochreiter and J{\"u}rgen Schmidhuber.
\newblock Long short-term memory.
\newblock {\em Neural Computation}, 9(8):1735--1780, 1997.

\bibitem{cho2014learning}
Kyunghyun Cho, Bart Van~Merri{\"e}nboer, Caglar Gulcehre, Dzmitry Bahdanau,
  Fethi Bougares, Holger Schwenk, and Yoshua Bengio.
\newblock Learning phrase representations using {RNN} encoder-decoder for
  statistical machine translation.
\newblock {\em arXiv preprint arXiv:1406.1078}, 2014.

\bibitem{ma2015long}
Xiaolei Ma, Zhimin Tao, Yinhai Wang, Haiyang Yu, and Yunpeng Wang.
\newblock Long short-term memory neural network for traffic speed prediction
  using remote microwave sensor data.
\newblock {\em Transportation Research Part C: Emerging Technologies},
  54:187--197, 2015.

\bibitem{fu2016using}
Rui Fu, Zuo Zhang, and Li~Li.
\newblock Using {LSTM} and {GRU} neural network methods for traffic flow
  prediction.
\newblock In {\em 31st Youth Academic Annual Conference of Chinese Association
  of Automation (YAC)}, pages 324--328. IEEE, 2016.

\bibitem{zhang2016dnn}
Junbo Zhang, Yu~Zheng, Dekang Qi, Ruiyuan Li, and Xiuwen Yi.
\newblock Dnn-based prediction model for spatio-temporal data.
\newblock In {\em Proceedings of the 24th ACM SIGSPATIAL International
  Conference on Advances in Geographic Information Systems}, pages 1--4, 2016.

\bibitem{yu2017spatiotemporal}
Haiyang Yu, Zhihai Wu, Shuqin Wang, Yunpeng Wang, and Xiaolei Ma.
\newblock Spatiotemporal recurrent convolutional networks for traffic
  prediction in transportation networks.
\newblock {\em Sensors}, 17(7):1501, 2017.

\bibitem{wu2018hybrid}
Yuankai Wu, Huachun Tan, Lingqiao Qin, Bin Ran, and Zhuxi Jiang.
\newblock A hybrid deep learning based traffic flow prediction method and its
  understanding.
\newblock {\em Transportation Research Part C: Emerging Technologies},
  90:166--180, 2018.

\bibitem{tak2016data}
Sehyun Tak, Soomin Woo, and Hwasoo Yeo.
\newblock Data-driven imputation method for traffic data in sectional units of
  road links.
\newblock {\em IEEE Transactions on Intelligent Transportation Systems},
  17(6):1762--1771, 2016.

\bibitem{bae2018missing}
Bumjoon Bae, Hyun Kim, Hyeonsup Lim, Yuandong Liu, Lee~D Han, and Phillip~B
  Freeze.
\newblock Missing data imputation for traffic flow speed using spatio-temporal
  cokriging.
\newblock {\em Transportation Research Part C: Emerging Technologies},
  88:124--139, 2018.

\bibitem{taylor2010convolutional}
Graham~W Taylor, Rob Fergus, Yann LeCun, and Christoph Bregler.
\newblock Convolutional learning of spatio-temporal features.
\newblock In {\em European Conference on Computer Vision}, pages 140--153.
  Springer, 2010.

\bibitem{fusco1995use}
Gaetano Fusco and Stefano Gori.
\newblock The use of artificial neural networks in advanced traveler
  information and traffic management systems.
\newblock In {\em Applications of Advanced Technologies in Transportation
  Engineering}, pages 341--345. ASCE, 1995.

\bibitem{chai2014root}
Tianfeng Chai and Roland~R Draxler.
\newblock Root mean square error ({RMSE}) or mean absolute error ({MAE})? --
  {A}rguments against avoiding {RMSE} in the literature.
\newblock {\em Geoscientific Model Development}, 7(3):1247--1250, 2014.

\bibitem{kingma2014adam}
Diederik~P Kingma and Jimmy Ba.
\newblock Adam: {A} method for stochastic optimization.
\newblock {\em arXiv preprint arXiv:1412.6980}, 2014.

\end{thebibliography}

\end{document}